\documentclass[letterpaper, 10 pt, conference]{ieeeconf}  % Comment this line out if you need a4paper

\IEEEoverridecommandlockouts                              % This command is only needed if 
\usepackage{biblatex}
\usepackage{hyperref}
\usepackage{multicol}
\usepackage{multirow}
\usepackage{xspace}
\usepackage{xcolor}
\usepackage{graphicx}
\usepackage{caption}
\usepackage{subcaption}
\usepackage{adjustbox}
\usepackage{xfrac}
\usepackage{listings}
\usepackage{float}
\usepackage{booktabs}
\usepackage{bm}
\usepackage{amsfonts}
\usepackage{tcolorbox}

\title{\LARGE \bf
ARCap: Collecting High-quality Human Demonstrations for Robot Learning with Augmented Reality Feedback
}

\author{Sirui Chen$^*$, Chen Wang$^*$, Kaden Nguyen, Li Fei-Fei, C. Karen Liu% <-this % stops a space
\thanks{$^*$ Equal contribution}
\thanks{Stanford University, Department of Computer Science}%
}

\begin{document}

\maketitle
\thispagestyle{empty}
\pagestyle{empty}

%%%%%%%%%%%%%%%%%%%%%%%%%%%%%%%%%%%%%%%%%%%%%%%%%%%%%%%%%%%%%%%%%%%%%%%%%%%%%%%%
\begin{abstract}
Recent progress in imitation learning from human demonstrations has shown promising results in teaching robots manipulation skills. To further scale up training datasets, recent works start to use portable data collection devices without the need for physical robot hardware. However, due to the absence of on-robot feedback during data collection, the data quality depends heavily on user expertise, and many devices are limited to specific robot embodiments. We propose ARCap, a portable data collection system that provides visual feedback through augmented reality (AR) and haptic warnings to guide users in collecting high-quality demonstrations. Through extensive user studies, we show that ARCap enables novice users to collect robot-executable data that matches robot kinematics and avoids collisions with the scenes. With data collected from ARCap, robots can perform challenging tasks, such as manipulation in cluttered environments and long-horizon cross-embodiment manipulation. ARCap is fully open-source and easy to calibrate; all components are built from off-the-shelf products. More details and results can be found on our website: \href{https://stanford-tml.github.io/ARCap}{stanford-tml.github.io/ARCap}

% Imitation learning has shown great progress in robot manipulation; collecting high-quality human demonstrations in the wild is essential for upscaling imitation learning for solving diverse tasks. Hindered by robots with different kinematics and collision geometry than humans, existing portable data collection systems require expertise in data collection and only work for a particular robot end-effector. We propose ARCap, a portable data collection system that provides visual, haptic feedback to guide users in collecting robot-executable demonstrations. Through user study, we show that ARCap allows users without any prior experience to collect data that matches robot kinematics and is free from collision. With data collected from ARCap, we show that robots can achieve non-trivial tasks such as manipulation under cluttered environments and long-horizon cross-embodiment manipulation with imitation learning. ARCap is fully open source and easy to calibrate; all components can be built from off-the-shelf products. More details can be found on our website: \href{https://stanford-tml.github.io/ARCap}{stanford-tml.github.io/ARCap}
\end{abstract}
%%%%%%%%%%%%%%%%%%%%%%%%%%%%%%%%%%%%%%%%%%%%%%%%%%%%%%%%%%%%%%%%%%%%%%%%%%%%%%%%

\section{Introduction}
% TODO: should add a teaser figure.
% Enable robot do every day task is important but challenging

Developing robots to assist with domestic tasks has the potential to enhance human quality of life and augment human capabilities. To achieve this, robots must be able to manipulate everyday objects in unstructured and often cluttered environments. Imitation learning using human demonstrations has made significant progress in recent years. Demonstration data collected via teleoperated robotic systems provide precise, in-domain observation and action pairs, enabling effective robot policy learning through supervised learning~\cite{aloha}. However, the requirement for a robotic system and a skilled human operator upfront significantly limits the accessibility and scalability of data collection.

Alternatively, human demonstrations can be collected using portable systems without the need for physical robot hardware~\cite{dobbe, chi2024universal, dexcap}. These systems leverage human dexterity and adaptability to directly manipulate objects in-the-wild, facilitating the creation of large-scale, diverse human demonstration datasets. However, due to the absence of robot hardware, whether the collected demonstrations are useful for training robot policies is not immediately apparent without going through a multi-step process. First, the differences in embodiment between humans and robots require data retargeting. Second, the retargeted data must be validated by replaying the motion on the actual robot interacting with real objects. Finally, the robot policy must be trained using validated data. The success of demonstrations critically depends on the demonstrator’s experience and awareness of the disparities between the robot and human geometry and capabilities. Failures can occur at the retargeting stage due to the robot's joint and speed limitations, during the validation stage due to incidental collisions, or at the policy training stage due to the mixture of invalid data.
\begin{figure}[t]
  \centering
  \includegraphics[width=\linewidth]{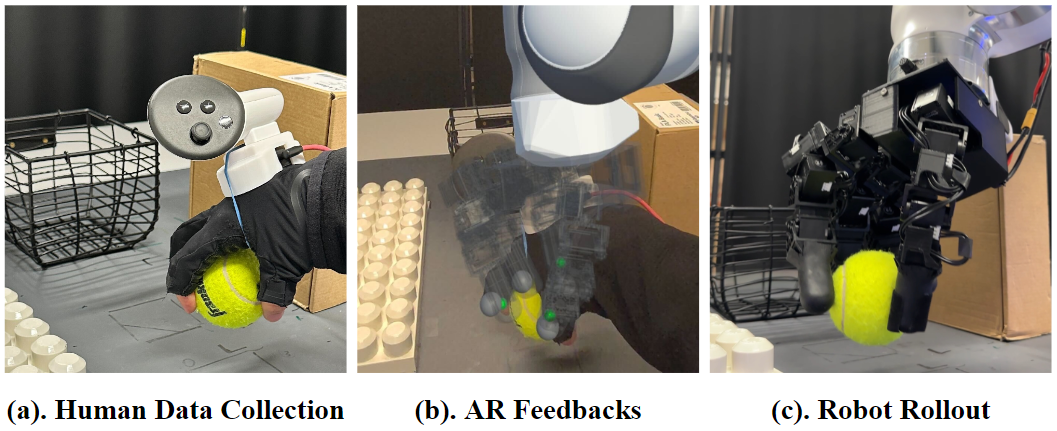}
  \caption{\textbf{ARCap System Overview.} (a) Collect human hand motion data. (b) Provide real-time AR feedback, visualizing a virtual robot retargeted to the human hand in AR display. (c) Rollout robot policies trained with the collected data.}
  \label{fig:teaser}
\end{figure}
% (2). virtual robot retargeted to the human hand and visualized in VR in realtime. Real robot learn from human demonstration.

This leads us to ask: Is there a way to inform users of potential failures during data collection so they can adjust and collect higher-quality data? One key observation from on-robot teleoperation is that when humans see incorrect robot motions, they quickly adjust the way of teleoperation to correct the error. This strong \textit{visual feedback} helps users collect data that is executable and suitable for the robot’s embodiment. Given the success of visual feedback in teleoperation, the question arises: Can we \textit{simulate} similar feedback in portable data collection systems to guide users in collecting high-quality demonstration data?

We propose ARCap, a novel data collection system that retargets and visualizes a robot's motion in real-time, providing the demonstrator with instant visual feedback during data collection and guiding them to collect robot-executable demonstrations. This is achieved by leveraging augmented reality (AR) technology, both as an interactive display and a powerful sensor that captures the user’s view of the environment. Using the AR display, we can simulate the robot's kinematics, overlay it in the headset, and provide visual cues for potential failure modes, such as exceeding the robot's joint or speed limits (e.g., the virtual robot fails to follow the human hand). Additionally, with recent advances in scene reconstruction in AR devices, we can perform collision checking between the virtual robot and the reconstructed environment. When a collision is detected, the system warns users with a blinking effect in the display and haptic vibration, prompting them to adjust their movements and leave enough space for the robot's embodiment. Besides improving data quality, ARCap can simulate any robot embodiment, enabling data collection for different robots (e.g., parallel-jaw grippers, multifinger dexterous hands). In our user study, we found that ARCap enables novice users without any prior data collection experience to collect high-quality data. These data are sufficient to train imitation learning policies, even for tasks like manipulation in cluttered environments—tasks that were impossible with previous systems lacking feedback. We also demonstrate that ARCap can collect data across different robot embodiments, enabling robots to accomplish challenging long-horizon manipulation tasks, such as stacking multilevel Lego towers.

\section{Related work}
\textbf{Learning from Demonstrations.}
Imitation Learning (IL) has proven effective in enabling robots to perform various manipulation tasks~\cite{calinon2010learning, 1014739, schaal1999imitation, kober2010imitation, englert2018learning, finn2017one, billard2008robot, argall2009survey}. While traditional IL methods like DMP and PrMP~\cite{schaal2006dynamic, kober2009learning, NIPS2013_e53a0a29, paraschos2018using} are highly sample-efficient, they face challenges in handling high-dimensional observation spaces. In contrast, recent IL approaches leveraging deep neural networks can learn policies directly from raw image inputs~\cite{mandlekar2021what, florence2019self, zhu2023viola}, even for complex robotic systems with bimanual manipulators~\cite{zhao2023learning, grannen2023stabilize, ace}. Although these methods are effective, scaling the amount of training data remains a significant hurdle. Teleoperation, a commonly used method for data collection in recent studies~\cite{zhang2018deep, florence2019self, mandlekar2019scaling, ke2020telemanipulation, wang2021generalization, zhu2022viola, brohan2022rt, wu2023gello, gao2024efficient, lin2024learning, aloha, mobilealoha, opentv, cp, he2024omnih2o,ding2024bunny, qin2023anyteleop, van2024puppeteer}. Many low-cost teleoperation systems built upon VR controller or hand tracking\cite{droid, opentv, ding2024bunny, van2024puppeteer} and master-slave joint mapping\cite{aloha, mobilealoha, alohaunleashed, ace, shawbimanual, fang2023low} were widely used. However, despite the low-cost nature of these action input devices, collecting data using teleoperation still requires the presence of a actual robot, which makes them expensive to distribute on a large scale. In contrast, our approach follows the recent fashion of collecting robot data without robot hardware~\cite{duan2023ar2, eve, umi, dexcap, dobbe}, allowing us to scale up the training data more efficiently. 

\textbf{Data Collection System without Robots.}
Collecting data in the wild without the presence of a robot and training robots with that data has become an attractive direction to lower the total cost of the system. Prior works such as \cite{umi, dexcap, dobbe} proposed low-cost, in-the-wild data collection systems. Compared to directly using human video for training\cite{videodex}, these systems capture more fine-grained human movement and have helped robots to achieve complex tasks such as tea preparation\cite{dexcap}, plate wiping\cite{umi, dexcap} and using air fryer\cite{dobbe}. Our ARCap system is another portable, in-the-wild data collection system; compared to existing systems, it provides visual, haptic feedback, which helps users without any data collection experience be aware of the embodiment gap between robots and humans. The most related work to ARCap is AR2-D2~\cite{duan2023ar2, eve}. ARCap, however, focuses on providing \textit{real-time} visual feedback and onboard collision checking using the reconstructed scene map. Additionaly, ARCap helps users collect data for different robot embodiments such as parallel-jaw grippers and multi-finger dexterous hands, by visualizing the retargeted robot in the AR display.

\section{Method}

ARCap is an AR-based data collection interface and policy learning framework designed to transfer human hand motion capture data to robot control policies. The main features of ARCap's system design are:
\begin{itemize}
    \item \textbf{Real-time feedback.} AR provides real-time visualization of the robot states, guiding users to collect high-quality and robot-reproducible demonstration data without physical robots.
    \item \textbf{Cross-embodiment.} AR visualization supports both parallel-jaw grippers and multifinger dexterous hands, allowing users to collect data for different types of robot hardware using the same system.
    \item \textbf{Portability.} With a self-contained power supply, storage, and wireless tracking, the system enables data collection in-the-wild.
\end{itemize}

In this section, we first describe the system design that enables these features, followed by the training policies for controlling real robots.

\subsection{ARCap System Design}

Recent advancements in portable robot data collection interfaces~\cite{dexcap, umi, dobbe} have made it possible to scale up robot data collection without needing a physical robot. However, since there is no real-time feedback from a robot during the data collection process, there is no guarantee that the collected data will be reproducible on an actual robot. Several failure modes have been observed: (1) Humans move too quickly for the robot to replicate; (2) Size differences between humans and robots cause the robot to collide with the environment, even when humans do not; (3) One data collection system is designed for one robot embodiment, requiring redesigns for different robot end-effectors. These observations begs the question: How can we alert humans about these issues during data acquisition and guide them to collect robot-reproducible data?

\textbf{Informative AR Feedbacks.}
In ARCap, we implement both visual and haptic feedback to inform users about camera visibility, robot kinematics, joint speed limits, and potential collisions between the robot and the environment.

\begin{figure}[t]
  \centering
  \includegraphics[width=\linewidth]{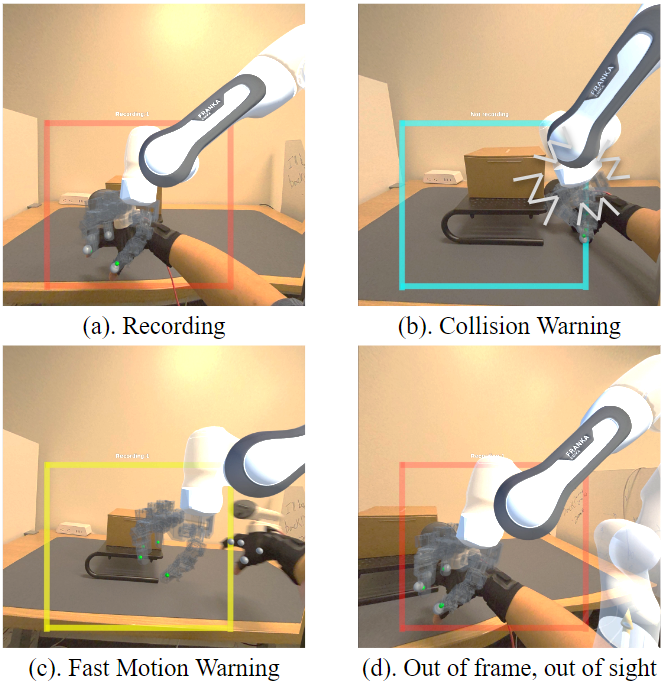}
  \caption{\textbf{Visualization of AR Feedback.} (a) Normal data recording: the red frame indicates visible region of the RGB-D camera. (b) Collision warning: when the virtual robot collides with the environment, the controller on the human gloves vibrates, and the frame blinks blue. (c) Fast motion warning: when the user moves faster than the robot's speed limits, the frame turns yellow. (d) Users can check if target objects are within camera's view during data collection.}
  \label{fig:interface}
  \vspace{-20pt}
\end{figure}

\paragraph{Real-time visibility checking}
One common failure mode for imitation learning is that the scene of manipulation is not always visible. This issue occurs frequently because RGB-D cameras used by robots usually have a narrower field of view compared to the cameras used for data collection---in our case, the passthrough cameras in Quest 3. To help the demonstrator always keep the manipulation scene within the field of view of the depth camera during data collection, we render a rectangular frame to visualize the actual field of view of the RGB-D camera, as shown in Fig.\ref{fig:interface}. When collecting data, users needs to actively keep the scene inside the frame to ensure visual data is being recorded properly.

\paragraph{Real-time retargeting}
When collecting data for a particular robot, the robot may have significantly different kinematics compared to the human arm and hand. To remind users about the kinematic limit, we rendered a virtual robot in AR and retargeted it to the user's hand. Different end-effectors may have different retargeting methods, as we will discuss in the next section. Before data collection, the user will place the virtual robot at a fixed location in the world frame. During data collection, the end-effector of the virtual robot will track the user's hand; whenever the user uses their hand to interact with an object in the scene, they need to consider whether the virtual robot could perform such an action. For example, for a virtual robot equipped with a parallel jaw gripper, if the user tries to reorient an object using finger gaiting, the action performed by the virtual robot will appear to be invalid, as shown in the attached video. As each joint of the robot arm has its speed limit, the virtual robot also implements such limits and won't exceed the speed limit to track the user's input. If the user moves their hand too fast, there will be a significant visual mismatch between the user's hand and the robot end-effector; the rectangular frame will also blink yellow to remind the user the robot has its speed limit.
% TODO: Add mismatch_action figure.
\paragraph{Real-time collision checking}
To remind users about the potential collisions between the robot and the environment, we also check the collision between the actual scene and the virtual robot. We found it is hard for humans to perceive depth accurately with passthrough cameras; only watching the movement of the virtual robot is not enough to avoid collision. We add extra haptic collision feedback when the virtual robot collides with the pre-scanned static scene by vibrating the mounted controller. The rectangular frame will also flash to provide a stronger collision feedback signal as shown in Fig.\ref{fig:interface}.

Using these real-time feedback signals, user can adjust their data collection strategies or delete the demonstration with severe constraint violations.

\textbf{Cross-Embodiement with One System.}
ARCap can visualize various end-effectors retargeted to the user's hand, enabling the collection of data for different robot embodiments without requiring hardware modifications. For any new robot embodiment, ARCap can be used for data collection as long as a retargeting process is in place that allows the robot to repeat human demonstrations. We present two real-time retargeting processes for different end-effectors attached to the Franka Panda arm: (1) Leap Hand, a fully actuated, four-finger dexterous hand, and (2) the Fin-ray gripper, a compliant parallel jaw gripper.
% Should add a figure shows retargeting
\begin{figure}[t]
  \centering
  \includegraphics[width=0.85\linewidth]{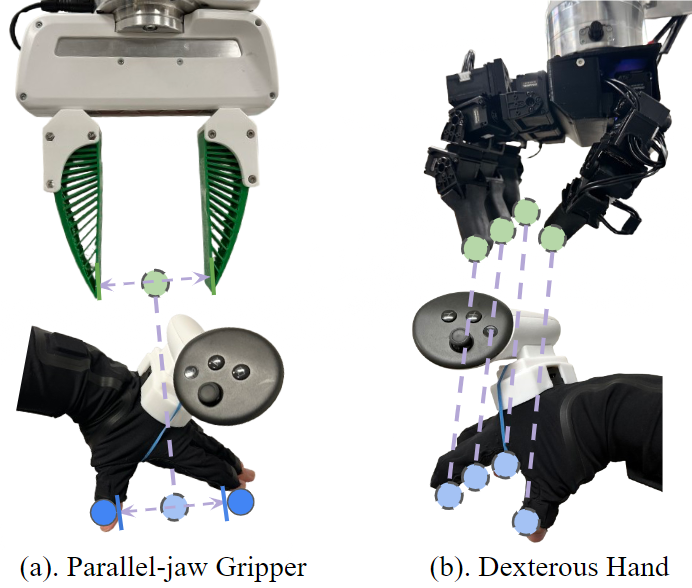}
  \caption{\textbf{Cross-Embodiment Data Collection.} (a) ARCap can collect data for parallel-jaw grippers by guiding the user to form their hands into a gripper-like shape. If the user changes the hand gesture, the retargeting error will be large. (b) For a multi-finger dexterous hand, ARCap retargets the robot's fingertips to match the human fingertips, with the robot's wrist orientation determined by the orientation of the controller mounted on the user's gloves.}
  \label{fig:retargeting}
  \vspace{-10pt}
\end{figure}
\begin{itemize}
\item \textbf{Dexterous hand.} Similar to \cite{dexcap}, we match the fingertips of a dexterous hand to the fingertips of a human in the world frame using inverse kinematics. The inverse kinematics problem is solved in two steps. It first solves the leap hand wrist pose to match the human wrist pose provided by the quest controller and then solves the robot fingertip positions to match human fingertip positions tracked by the Rokoko data glove. As each finger of the leap hand has one redundant degree of freedom, we need to add null space regulation to encourage a natural hand posture and avoid self-collision between fingers. We use null space IK solver from Pybullet\cite{pybullet}, which solves the current joint angle based on the previous solution in real-time.
\item \textbf{Parallel jaw gripper.} For the parallel jaw gripper, the user mimics it by using their index finger and thumb. As shown in Fig.\ref{fig:retargeting}, the midpoint of the gripper tips is aligned with the midpoint between the user's index finger and thumb, while the wrist orientation tracked by the controller sets the gripper's orientation. Since the gripper can only fully open or close, its state is determined by the distance between the user's index finger and thumb. If the distance is greater than gripper width at open state, it is set to open; otherwise, it is set to close. On real robots, the gripper responds to open and close commands at 1Hz. In our retargeting process, if the user opens and closes their hand too frequently, the virtual gripper won't open or close till 1s from changing to the previous state.
\end{itemize}

\begin{figure}[t]
  \centering
  \includegraphics[width=0.85\linewidth]{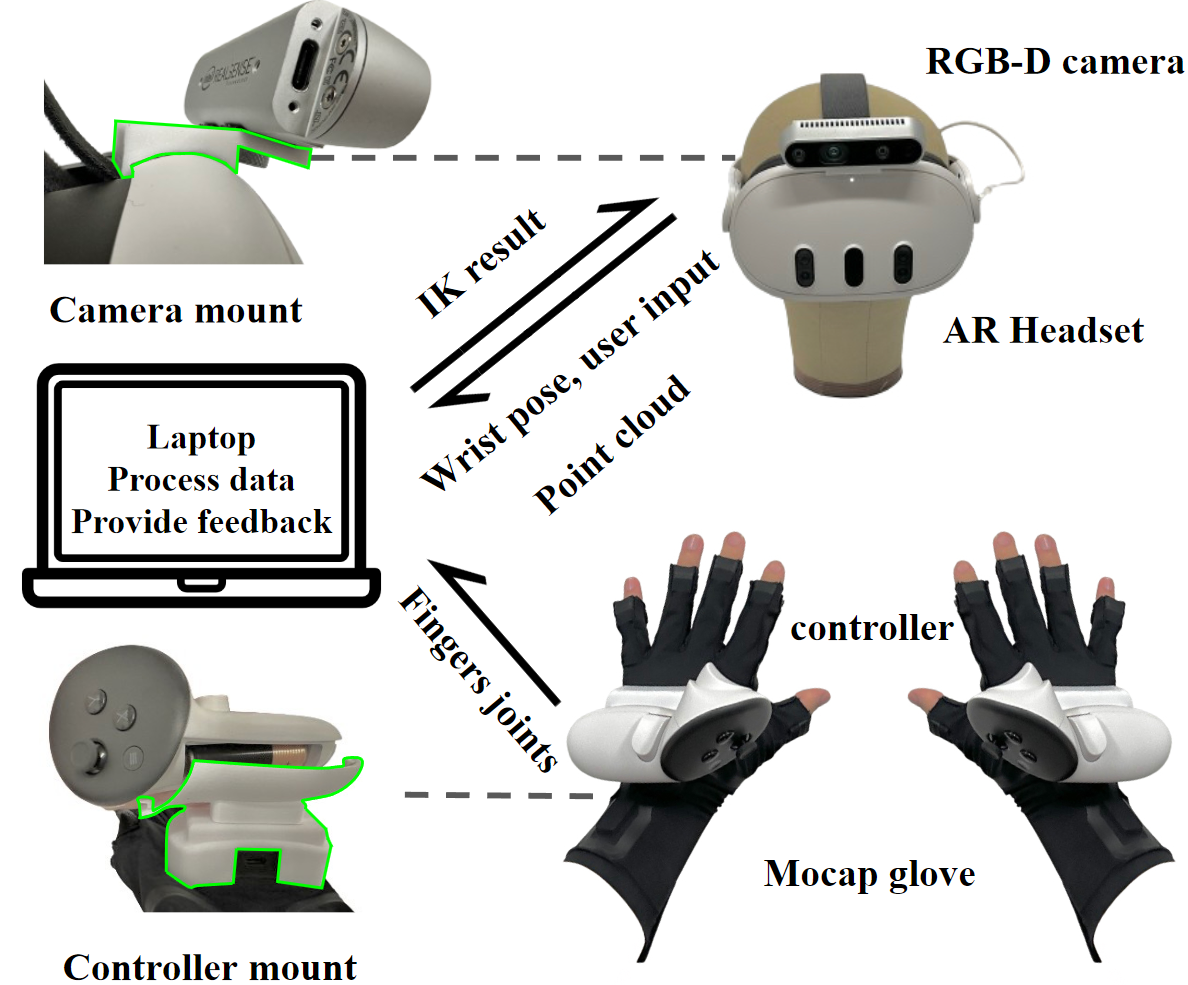}
  \caption{\textbf{ARCap System Layout.} The user wears an AR headset and motion capture gloves, with controllers mounted on the gloves for tracking the 6D pose of the palms. Data is stored on a laptop carried in the backpack.}
  \label{fig:system}
  \vspace{-20pt}
\end{figure}

\textbf{Portable and Reproducible Design.} ARCap is designed to be a low-cost, portable system that is easy to reproduce and calibrate, while accurately capturing detailed hand motions. It also ensures user comfort during various tasks with minimal obstruction. To achieve these goals, ARCap is built around the Meta Quest 3 VR headset, as shown in Fig.\ref{fig:system}. The headset serves as both a display for feedback and a sensor hub, providing spatial tracking for itself and both controllers.
A RealSense D435 camera is mounted on top of the headset using a 3D-printed bracket to capture 3D visual information, which is stored as point clouds. Since accessing the internal Quest 3 camera is difficult, future versions of ARCap could leverage the built-in RGB-D camera of an AR headset. 

For wrist and hand motion capture, Quest 3 controllers are attached to the top of Rokoko data gloves. The controllers track wrist position and orientation relative to the headset, while the data gloves capture fingertip positions relative to the wrist. Using the headset’s built-in SLAM function, we can access both visual and motion data within a world frame.

Calibrating the system can be time-consuming due to the need to fine-tune relative transformations between components. To streamline this process, the camera is mounted directly to the headset, and the controllers are attached to the gloves with unique 3D-printed mounts, allowing future ARCap setups to reuse the same calibration parameters. A laptop is connected to process visual data and provide additional storage for collected data. Like DexCap \cite{dexcap}, ARCap system is portable and can be carried in a backpack, enabling data collection without external infrastructure.

\subsection{Imitation learning}
% What data is recorded?
% Should briefly recap dexil and input and action
\subsubsection{Data processing}
ARCap records the following data:
\begin{itemize}
\item Colored point cloud in the camera frame
\item Joint angle for the virtual robot solved by IK
\item Headset pose in the world frame
\item Virtual robot pose in the world frame
\end{itemize}
The collected data can be used for imitation learning with a simple post-processing procedure. We first transform every data into the world frame. For point clouds, we further crop them in the world frame to remove background objects and the desktop. In the collected data, the hand and arm of a human user are visible. To reduce the visual gap, we superimpose a point cloud of the virtual robot visible by the depth camera in our point cloud dataset. After processing, all data for a single task will be stored in one hdf5 file. 
\subsubsection{Training and testing} % More details
\begin{figure}[t]
  \centering
  \includegraphics[width=0.85\linewidth]{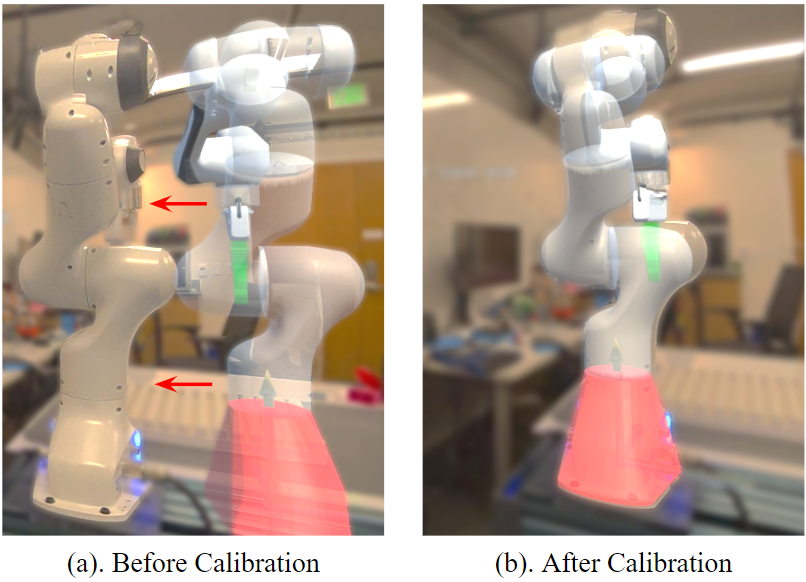}
  \caption{\textbf{AR-based Camera Calibration.} When calibrating the camera, users align the virtual robot's base with the actual robot's base.}
  \label{fig:calibration}
  \vspace{-20pt}
\end{figure}
With processed data, we use diffusion policy for imitation learning. For encoding 3D point cloud, similar to \cite{dexcap, 3dp}, a simple point net is used to compress colored point clouds into a latent vector. After that, the latent vector is concatenated with the current joint angle of the robot arm and hand as observation $\bm{o}$. The generated action $\bm{a}$ consists of the target joint angles of both robot arm and hand; for dex hand, $\bm{a}$ consists of the target joint angles of each finger; for parallel jaw gripper, $\bm{a}$ include a binary open and close command. Our training and testing pipeline is built upon robomimic\cite{robomimic}, a unified framework for robot imitation learning. When testing trained policy, we can utilize the ARCap system to simplify the hand-eye calibration process. As shown in Fig.\ref{fig:calibration}, to compute the camera pose related to the robot base, we align the base of the virtual robot to the base of the actual robot in the ARCap application.

\begin{figure*}[!ht]
  \centering
  \includegraphics[width=0.95\linewidth]{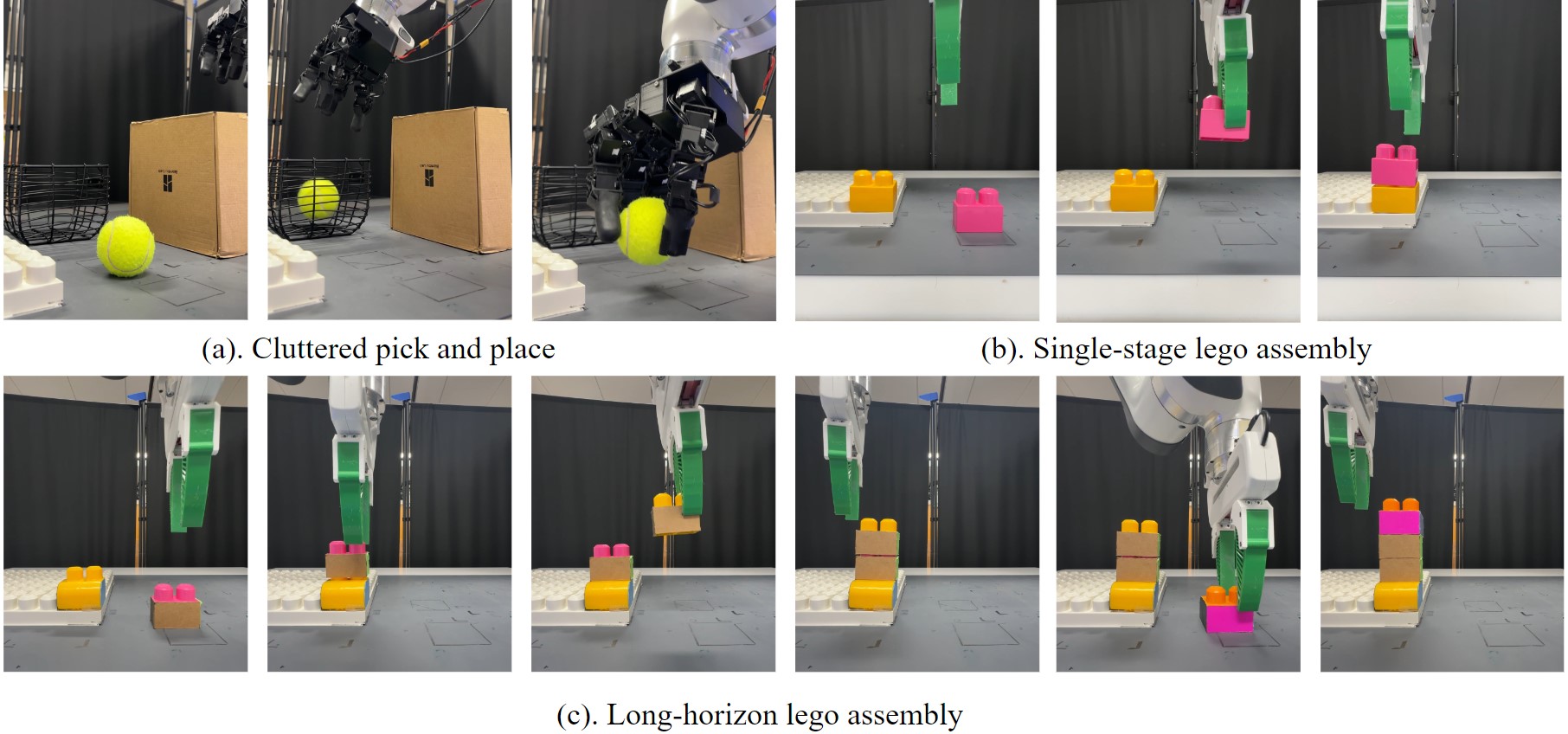}
  \caption{\textbf{Experiment Tasks.} (a) Pick-and-place in a cluttered scene with a dexterous robot hand. (b) Single-stage Lego assembly with a parallel-jaw gripper. (c) Long-horizon assembly of three Lego blocks.}
  \label{fig:tasks}
  \vspace{-15pt}
\end{figure*}
\section{Experiments}
% Hypothesis:
% General user can collect more usable data with ARCap
% With ARCap human collected data can enable manipulation under cluttered environment
% With ARCap human collected data can enable manipulation with different embodiment.
% With ARCap we can scale up to long horizon collaborative tasks (maybe)
We design experiments to answer the following questions:
\begin{itemize}
\item[\textbf{Q1}] Does ARCap enable general users to collect higher-quality data
\item[\textbf{Q2}] Can data collected by ARCap help robots to manipulate under a cluttered environment?
\item[\textbf{Q3}] Can data collected by ARCap work on the robots with significantly different embodiments?
\item[\textbf{Q4}] Does data from ARCap good enough for achieving long-horizon manipulation?
\end{itemize}
\begin{figure}[!h]
  \centering
  \includegraphics[width=\linewidth]{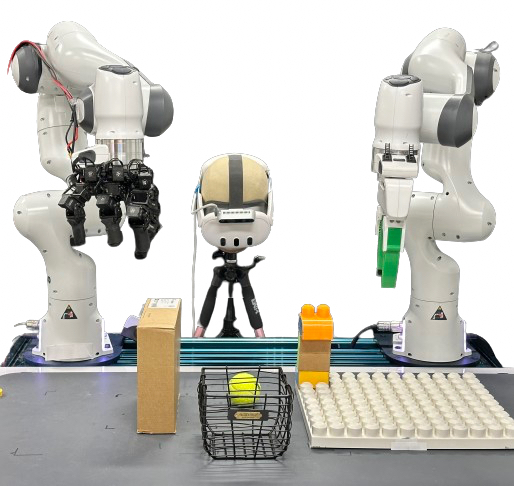}
  \caption{\textbf{Setups for Real Robot Evaluation.} During evaluation, we mounted the headset on a tripod and connected its camera to the robot workstation. The trained policy takes point cloud observations from the camera attached to the headset and outputs actions to control the robots.}
  \label{fig:setup}
  \vspace{-10pt}
\end{figure}
\begin{figure*}[!ht]
  \centering
  \includegraphics[width=\linewidth]{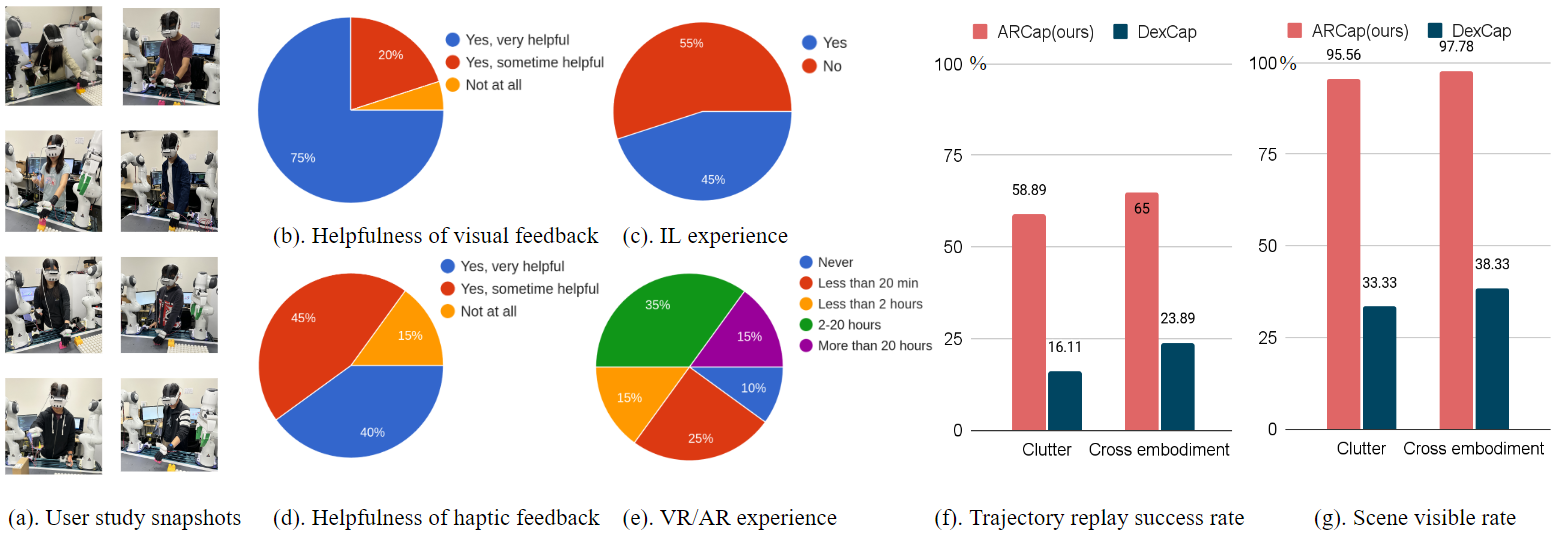}
  \caption{\textbf{Results of User Study.} (a) We invited 20 users with different backgrounds to collect data with DexCap~\cite{dexcap} and ARCap. (b)-(e) Survey results of user experience. (f)-(g) On-robot evaluation results with collected data.}
  \label{fig:user_study}
  \vspace{-15pt}
\end{figure*}
\subsection{Experiment setup}
We used two Franka Panda arms in our experiment, one attached with a Leap hand and another one attached with a Fin-ray gripper. Two robots shared the same workspace. For data collection, a Quest 3 headset runs a Unity application for visualization and data streaming, and a Windows laptop with an i5-13200H CPU is used for solving IK and storing data. For training and testing autonomous policy, we use a workstation with a single RTX3090 GPU and i7-13700 CPU. When testing, we calibrate the camera using the above-mentioned process and put the headset on a dummy head to serve as an RGB-D camera, as shown in Fig.\ref{fig:setup}.
\subsection{User study}

% Should I provide a detailed breakdown of failure mode?
To answer the \textbf{Q1}, we conduct a user study and invite 20 participants to use our new system ARCap with visual haptic feedback and the previous system DexCap, which has no feedback. Users have different exposure to VR/AR devices, and half of them have no data collection or robot learning experience; Fig.\ref {fig:user_study}.(c,e) shows the demographic of our participants. Moreover, none of the participants used either ARCap or DexCap before participating in this study. 

Test participants are asked to collect data for two tasks as shown in Fig.\ref{fig:tasks}: (1). Picking and placing a tennis ball with obstacles using a dexterous leap hand. (2). Assembling a single Lego block with a fin-ray parallel jaw gripper. The first task aims to test whether feedback from ARCap can help the user avoid collision under a cluttered environment; the second task aims to test whether feedback can help the user collect valid data under different end-effector embodiment. Each task has 3 initial states, and the subject needs to collect 3 trajectories on each initial state.

From experience with training imitation learning policies, trajectory reproducibility and scene visibility are two essential factors determining the quality of collected demonstrations. In our user study, to quantitatively measure the quality of collected data, we test whether robots can replay the collected trajectory and accomplish the same task, as well as whether the manipulation scene is always visible during the data collection process. 

Fig.\ref{fig:user_study} shows the replay success rate and scene visible rate of both tasks; ARCap achieve over 40\% higher replay success rate and over 60\% higher scene visible rate compared to DexCap. In our evaluation, we found that ARCap frequently prevents failures caused by collisions or kinematic limits in both tasks. It can also significantly reduce failure cases caused by users ignoring the gripper closing speed limit. A post-survey was conducted by the participants and summarized in Fig.\ref{fig:user_study}.(b,d). Results show that most participants found visual and haptic feedback useful in helping them improve data collection strategies.

\subsection{Manipulation in cluttered environment}
To verify whether data collected by ARCap can actually help robot imitation learning to achieve manipulation in a cluttered environment. We collected two 30-minute datasets using both systems and trained two diffusion policies on each of them. These two datasets are collected by the authors of this paper, who are familiar enough with both ARCap and DexCap. After training, we evaluate the policy using 20 trials with different initialization. Shown in Tab.\ref{tab:result}, ARCap can achieve a 35\% higher success rate compared with DexCap, and no collision ever happens when testing ARCap policy.  We also merge 30-minute data crowd-sourced from multiple first-time users during user study and train an autonomous policy from them. ARCap policy can achieve a 60\% success rate across 3 designated initial states, while DexCap policy failed everytime across different trials, shown in Tab.\ref{tab:result}

\begin{table}[h]
\begin{tabular}{l|cccc}
\toprule
\textbf{Cluttered pick and place} & \multicolumn{2}{c}{Expert}       & \multicolumn{2}{c}{User}         \\ \hline
DexCap~\cite{dexcap}                            & \multicolumn{2}{c}{0.25}         & \multicolumn{2}{c}{0}            \\ 
ARCap                             & \multicolumn{2}{c}{\textbf{0.7}} & \multicolumn{2}{c}{\textbf{0.6}} \\ \hline
\textbf{Long-horizon lego assembly}    & Stage 1         & Stage 2        & Stage 3         & Full           \\ \hline
DexCap~\cite{dexcap}                            & 0.5               & 0.15              & 0.05               & 0.0    \\
ARCap                             & \textbf{0.7}    & \textbf{0.8}   & \textbf{0.85}   & \textbf{0.4}   \\
\bottomrule
\end{tabular}
\caption{Success Rate of Autonomous Policy}
\label{tab:result}
\vspace{-20pt}
\end{table}

\subsection{Long horizon manipulation with different embodiments}
To answer \textbf{Q3} and \textbf{Q4}, we also show that ARCap can collect high-quality data with embodiment significantly differ than human and help robots achieve the task using imitation learning. We demonstrate this capability using a long-horizon, three-stage Lego assembly tasks with parallel jaw gripper as demonstrated in Fig.\ref{fig:tasks}.(c). This task is challenging as it requires policy to learn different grasp and assembly actions for Lego blocks at different stages. We used both DexCap and ARCap to collect two datasets of one-hour human manipulation and trained two policies based on them. We first evaluate the success rate independently at different stages. In this evaluation, humans reset the Lego tower to one stage prior to assembly after each trial.  As shown in Tab.\ref{tab:result}, ARCap can achieve 70\%, 80\%, and 85\% success rates at stages 1, 2, and 3. We also evaluate the policy on assembling all 3 stages autonomously in which humans always reset the Lego tower to the base level; ARCap policy achieves a 40\% success rate in full autonomous assembly, which is, on average, 51\% higher than the DexCap policy. The policy could also react at different stages when humans disassemble the Lego tower, as shown in our supplementary video.
% Should state challenges: finger matching, delay, binary action
% For result should also report success rate at different stages and overall success rate.

%%%%%%%%%%%%%%%%%%%%%%%%%%%%%%%%%%%%%%%%%%%%%%%%%%%%%%%%%%%%%%%%%%%%%%%%%%%%%%%%
\section{Conclusions and Future Work}
We propose ARCap, a portable data collection system that allows users without prior experience to collect high quality data across different embodiments via visual, haptic feedback. Using ARCap, we can teach robot manipulation in cluttered environments and achieve horizon cross embodiment manipulation with imitation learning. In the future, with additional design in feedback and retargeting process, ARCap also record human torso movement to collect data for mobile robots or humanoids. Currently, user improves their data collection strategies passively from feedback; with VLM, ARCap could also provide instruction for users to actively improve their data collection strategies and efficiency.


%%%%%%%%%%%%%%%%%%%%%%%%%%%%%%%%%%%%%%%%%%%%%%%%%%%%%%%%%%%%%%%%%%%%%%%%%%%%%%%%
% \section*{ACKNOWLEDGMENT}

% We thank Weizhuo Wang and Haochen Shi for their suggestions for the mount design. We also thank Robert Wang, Joao Araujo, Yunhao Liu, and Jared Bienz for the insightful discussion regarding Quest 3 app development and sensor access. This work is supported by xxx \eric{add funding info}

%%%%%%%%%%%%%%%%%%%%%%%%%%%%%%%%%%%%%%%%%%%%%%%%%%%%%%%%%%%%%%%%%%%%%%%%%%%%%%%%
\printbibliography

\end{document}